\documentclass[runningheads,a4paper]{llncs}

\usepackage{amssymb}
\usepackage{graphicx}
\usepackage{url}
\newcommand{\keywords}[1]{\par\addvspace\baselineskip
\noindent\keywordname\enspace\ignorespaces#1}

\usepackage{subfig}
\usepackage{hyperref}

\usepackage{tabularx}
\usepackage{pbox}

\usepackage{listings,array,varwidth}

\newcommand{\textincon}[1]{%
{\fontfamily{fi4}\selectfont #1}}

\begin{document}

\mainmatter  % start of an individual contribution

\title{Text-based LSTM networks for\\Automatic Music Composition}

\titlerunning{Text-based LSTM networks for Automatic Music Composition}

%\author{Authors\and AuthorB\inst{2} \thanks{This work is supported by **********\textit{(will be filled in the camera-ready version)}.}}
\author{Keunwoo Choi \and George Fazekas \and Mark Sandler \thanks{This paper has been supported by EPSRC Grant EP/L019981/1, Fusing Audio and Semantic Technologies for Intelligent Music Production and Consumption.
}}

%
% if the names of the authors are too long for the running head, please use the format: AuthorA et al.
% \authorrunning{Keunwoo Choi et al.}

\institute{The Centre for Digital Music, Queen Mary University of London \\ \email{\{keunwoo.choi, g.fazekas, mark.sandler\}@qmul.ac.uk} }


\maketitle

\begin{abstract}
In this paper, we introduce new methods and discuss results 
of text-based LSTM (Long Short-Term Memory) networks for 
automatic music composition. The proposed network is designed 
to learn relationships within text documents that represent 
chord progressions and drum tracks in two case studies. In 
the experiments, word-RNNs (Recurrent Neural Networks) 
show good results for both cases, while character-based RNNs (char-RNNs) 
only succeed to learn chord progressions. The proposed system 
can be used for fully automatic composition or as semi-automatic 
systems that help humans to compose music by controlling a 
diversity parameter of the model. 

% \emph{abstract} 
\end{abstract}

\section{Introduction}\label{sec:introduction}
Music composition is considered creative, intuitive and therefore inherently human. 
Nevertheless, it has a long history of mathematical approaches since Hiller 
and Isaacson proposed to use \textit{Markov} chains for automatic 
composition \cite{hiller1959experimental}. The field of automatic composition 
includes a wide range of tasks such as the composition of melody, chord, 
rhythm \cite{kleedorfer2008oh}, and even lyrics \cite{de2002ai}, i.e. every typical 
components of music, and has been subject to numerous research studies. 
There are many applications for automatic composition too; automatic 
background music generation, AI-assisted 
composition systems and 
improviser software\footnote{\url{http://jukedeck.com}, \url{http://arpegemusic.com}, \href{https://www.pgmusic.com}{\textit{Band-in-a-Box, PG Music Inc.}}} for example.

Music can be represented as a sequence of events and thus it can be 
modelled as conditional probabilities between musical events. For 
example, in harmonic tracks, some chords are more likely to occur 
than others given the previous chords, while the whole chord progressions often depend 
on the global key of the music. In many automatic composition systems, 
these relationships are simplified by assuming that the probability of
the current state $p(n)$ only depends on the probabilities of the states 
in the past $p(n-k)...p(n-1)$. A sequence of musical events - notes, chords, 
rhythm patterns - is generated by predicting the following event given a seed sequence.

{\em Hidden-Markov models (HMMs)} are one of the most popular methods to model and predict sequences. HMMs are based on the assumption of $k=1$ (Markov assumption) given the sequence of the hidden states which determine the visible states. Choral harmonisation is generated after learning chorales by Bach using a HMM in \cite{allan2005harmonising}, where $229$ and $153$ chorales are used for training and testing, respectively. In \cite{simon2008mysong}, chord progressions are generated to accompany a melody to help non-musicians to create music using a HMM. The training set of the HMM consists of $298$ lead sheets including pop, rock, R\&B, jazz, and country music. In the prediction, the system generates chords using a $62$$\times$$62$ chord transition probability matrix. In practice, HMMs had been the most suitable for time-series modelling given the data, computing power, and feasible optimisation strategies. One of the drawbacks of HMMs, however, is the inefficiency of {\em 1-of-K} scheme of its hidden states. The memory of HMM is limited to $log_2(N)$ bits when there is $N$ hidden states, which requires to learn $N^2$ parameters for the transition matrix.

{\em Recurrent Neural Networks} ({\em RNNs}) allow for incorporating long term dependency in the model. {\em Jordan net} \cite{jordan1986attractor}, a simple version of RNNs, is used in \cite{lewis1989algorithms} to generate chord sequences. In \cite{mozer1994neural}, melodies were generated by a system named \textit{CONCERT}, which is trained on sets of $10$ Bach pieces to generate melodies by note-wise prediction. One ability \textit{CONCERT} lacks is to learn the global structure; this may be due to the difficulty of training an RNNs. 
Theoretically, it can remember infinitely long sequences, although in practice it is limited by the \textit{vanishing gradient} problem \cite{hochreiter2001gradient}. During the training of back-propagation through time, the gradient is extremely diminished by multiplications of sigmoid operations. 

{\em LSTM (Long Short-Term Memory)} units solved this vanishing gradient problem \cite{hochreiter2001gradient}. LSTM allows the gradient to be flowed by a separate path with not multiplication but \textit{addition} operations. LSTM is adopted in \cite{eck2002first} to learn $12$-bar Blues chords progressions and melodies. \cite{lambert2015perceiving} focuses on the generation of percussive tracks using LSTM network. The network in \cite{lambert2015perceiving} directly analyses audio content of drum tracks and learns features using LSTM.

In this paper, we introduce applications of character- and word-based RNNs with LSTM units for the automatic generation of jazz chord progressions and rock music drum tracks. Our work is differentiated from previous works by two aspects. First, the LSTM networks we use are designed to learn from text data rather than representations of musical symbols or numeric values. Directly using text data minimises the overall design procedures for the encoding-decoding scheme and the network. Second, compared to the previous research \cite{allan2005harmonising},\cite{eck2002first},\cite{simon2008mysong}, the LSTM networks is trained using a large dataset, which enables itself to learn more complex relationship between the chords in a large set.

In the Section \ref{sec:arch}, we introduce character-based RNNs and the proposed architecture. In Sections \ref{sec:chord} and \ref{sec:rhythm}, two case studies on the applications of RNNs to automatic composition are explained - for jazz chord progressions and rock music drum tracks. We conclude the work in Section \ref{sec:con}.

\section{The architecture}\label{sec:arch}
\subsection{Character-based RNNs}
% The LSTM unit is a variant of the unit of RNNs and proposed in \cite{hochreiter2001gradient}. Unlike conventional RNNs, a RNNs with LSTM units (LSTM networks) selectively inputs, outputs, and forgets using the corresponding gates which enables the LSTM networks to be longer than conventional RNNs are.%\footnote{By username:\textit{BiObserver}, CC BY-SA 4.0, \\https://commons.wikimedia.org/w/index.php?curid=43992484} 

{\em Char-RNNs} are RNNs with character-based learning \cite{sutskever2011generating}, which is different from the conventional approach of word-based learning. When applied to the texts of chords, a char-RNN predict a vector that corresponds to a character (e.g. predict \textit{a} based on \textit{C:m}, and predict \textit{j} based on \textit{C:ma}), while a word-RNN predicts a vector, which corresponds to a unique chord (e.g. \textit{C:maj} based on \textit{G:maj}).
Using char-RNNs in this work has two merits. 

First, it is based on the minimal assumption - there is no constraint on the form of the text representation of music. It is worth inspecting if RNNs can learn musical information with such a weak assumption. 

Second, fewer number of characters means fewer number of states, which results in reducing the computational cost. From a linguistics point of view, sequence learning methods such as HMMs and RNNs used to model each {\em word}(e.g. chord) as a {\em state} as it is natural to find the relationships between words. One drawback of word-based learning is the large number of states (or the size of vocabulary); in natural language processing tasks, the vocabulary size easily exceeds few thousands to even few millions. In the proposed method the size of the chord vocabulary is $1$,$259$. With character-based prediction, this decreases to $39$. 

The price of small vocabulary size is a longer sequence; as we need to learn character by character, the model should \textit{remember} a longer sequence of states. As mentioned above, the LSTM unit helps the RNNs to learn this long-term dependency better. This trade-off does not necessarily benefit as in Section \ref{sec:rhythm}.

% \begin{figure}[t]
% \centering
% \includegraphics[width=0.7\columnwidth]{lstm_unit.png}
% \caption{A diagram of an LSTM unit\label{fig:lstm}}
% \end{figure}

\subsection{The Proposed Architecture}\label{sec:proposed}
% {\em required}\\
% (\texttt{http://icmc2016.unt.edu/}).
%------------------------------------------------------%
% \section{The Proposed Architecture}\label{sec:method}

%\begin{figure}[h]
%\centering
%\includegraphics[width=0.18\columnwidth]{CSMC2016-diagram-lstms.pdf}
%\caption{The block diagram of the networks \label{fig:diagram}}
%\end{figure}

%Figure \ref{fig:diagram} shows the block diagram of the LSTM networks we use in the experiments. 
We use two LSTM layers, each of which consists of 512 hidden units. Dropout of $0.2$ is added after every LSTM layers \cite{zaremba2014recurrent}.

We use the {\em Keras} deep learning framework \cite{chollet2015}. During the optimisation, categorical cross-entropy is used as a loss function and optimisation is performed by ADAM \cite{DBLP:journals/corr/KingmaB14}. This optimiser shows an equivalent final performance to Stochastic Gradient Descent with Nestrov momentum with faster convergence.

The prediction is stochastic. In each prediction for time index $n$, the network outputs the probabilities of every states. To make the system \textit{tunable}, We employ a diversity parameter $\alpha$ in the prediction stage (see Eqn. \ref{eq:diversity}), which suppresses ($\alpha<1$) or encourages ($\alpha>1$) the diversity of prediction by re-weighting the probabilities. In detail, the probabilities of $i$-th state, $p_i$, are re-weighted as $  \hat{p_i} = \exp{(\log(p_i)/\alpha)}$. Then, one of the states is selected by sampling a state according to the re-weighted probabilities.

As stated in Section \ref{sec:chord}, we perform experiments with char- and words-RNNs. We keep the same size and number of layers for both networks, although they result in different effective lengths; for example, manifold states are needed to be predicted to complete a chord in char-RNNs while each state correspond to a chord in word-RNNs.

The dataset, code and audio files are released on web.\footnote{\url{https://github.com/keunwoochoi/lstm_real_book}\\ \url{https://github.com/keunwoochoi/LSTMetallica} \\ \url{https://soundcloud.com/kchoi-research/sets/lstm-realbook-1-5} \\ \url{https://soundcloud.com/kchoi-research/sets/lstmetallica-drums}}
% Three-layer, LSTM of 512 units, 0.2 dropout for every, and softmax. loss with categorical_crossentropy and ADAM optimiser on Keras.

%%%%%%%%%%%%%%%%%%%%%%%%%%%%%%%%
% CASE 1. CHORDS
%%%%%%%%%%%%%%%%%%%%%%%%%%%%%%%%
\section{Case Study 1: Chord progressions}\label{sec:chord}
\subsection{Representation}
The goal of this experiment is to generate chord progressions by training an LSTM network on jazz chord progressions. Here, we do not use any musical interpretation of the chords such as binary vectors to represent pitch and chords (as in \cite{franklin2006recurrent}) but completely rely on their text representations. Table \ref{table:chord_texts} shows an example of a chord progression and the corresponding texts. The left is an example of a chord notation in The Realbook score, where the positions of chords are loosely related to the timings of chord changes. The score on the left is converted into the text on the right, which specifies every chord for each quarter note.

\begin{table}[t]
\begin{center}
\begin{tabular}{|c|c|}
\hline
\parbox{3.1cm}{
  \vspace{.2\baselineskip}
  \textincon{F:9  \hphantom{ MMM M MM}    $|$\\ D:min7 \hphantom{MM} G:9 \hphantom{M}$|$\\ C:maj \hphantom{MaM} F:9  \hphantom{al}  $|$\\ C:maj \hphantom{MMMMl M} $|$} \vspace{.05\baselineskip}} &
\parbox{4.5cm}{
  \vspace{.2\baselineskip}
  \textincon{\small{F:9 F:9 F:9 F:9 D:min7 D:min7 G:9 G:9 C:maj C:maj F:9 F:9 C:maj C:maj C:maj C:maj}}
  \vspace{.05\baselineskip}
  }
  \\
\hline
\end{tabular}
  \vspace{.05\baselineskip}
  
\caption{An example of the text representations of chord progressions in score (left) and the training data (right). A 4-bar chord progression is generally written in the form on the left, where the positions of the chords loosely indicate the chord change timings. 
%(note that the scores in the training dataset clearly specify the timings of chords)
On the right, the text show how the score on the left is represented in the training data. Here, the chords for every quarter notes are explicitly written and bar indicators are removed.}
\label{table:chord_texts}	
\end{center}
\end{table}
\vspace{0cm}

We used $2,486$ scores from The Realbooks and The Fakebooks as training data. Every score file was parsed from \textit{band-in-a-box} format to \textit{.xlab} format. Then they were transposed to the key of C while every blank quarter note was filled with its preceding chord as in the Table \ref{table:chord_texts}. Finally, we put \texttt{\char`_START\char`_} and \texttt{\char`_END\char`_} flags (any distinctive words can be used as flags) at the beginning and the end of each score. 

Although the key was transposed to C, only $867$ (out of $2$,$846$) scores end with \textincon{C:maj} ($30\%$), followed by $489$ \textincon{G:7} (17\%), $186$ \textincon{C:maj6} ($7\%$), $52$ \textincon{F:maj} ($2\%$), and $1$,$252$ scores end with the others -- $237$ chords ($46\%$). This is because the The Realbook chord progressions usually end with chords for a \textit{turn-around} to make the progressions natural to repeat the score.

There were $1,259$ unique chords in the training dataset. In other words, the vocabulary size of word-RNN was $1,259$. However there were only $39$ characters in total, which significantly reduced the computation of char-RNN. The total numbers of chords (words) and characters were $539,609$ and $3,531,261$, respectively.

\subsection{Results}
We set the system to output a chord progression for every diversity parameter $\alpha$ after every iteration. In this paper, we present four results from each networks (char-RNNs and word-RNNs), part of which are reported in the Table \ref{table:chord_results}. For simplicity, we added bar symbols $|$ and removed repeating chords in the same bar, e.g. \textincon{$|$ C:7 C:7 C:7 C:7 $|$} reduced to \textincon{$|$ C:7 $|$} and \textincon{$|$ C:7 C:7 E:min E:min $|$} reduced to \textincon{$|$ C:7 E:min $|$}.

\begin{table}[b!]
\begin{center}
\begin{tabular}{|c|c|c|}
\hline %  ROW 0
\parbox{0.6cm}{
  \vspace{.2\baselineskip}
	$i$
  \vspace{.2\baselineskip}
  } &
  \parbox{0.6cm}{
  \vspace{.2\baselineskip}
  $\alpha$
  \vspace{.2\baselineskip}
  } &
\parbox{10.0cm}{
  \vspace{.2\baselineskip}
  \textbf{Chord progressions}
  \vspace{.2\baselineskip}
  }
  \\
  \hline %  ROW 1
\parbox{0.6cm}{ 1 } &
\parbox{0.6cm}{ 0.8} &
\parbox{10.0cm}{
  \vspace{.2\baselineskip}
  \textincon{\small{C:maj $|$ G:7 $|$ ... $|$ G:7 $|$ C:maj}}
  \vspace{.2\baselineskip}
  }
  \\
  \hline % ROW 2
\parbox{0.6cm}{ 1 } &
\parbox{0.6cm}{ 1.2} &
\parbox{10.0cm}{
  \vspace{.2\baselineskip}
  \textincon{\small{A\#:maj $|$ A:7 $|$ A:7 D:min7 D:min7 D:min7 $|$ D:hdim C:hdim $|$C:hdim $|$ C:hdim G:9 G:9 D:min7 $|$ D:min7 D\#:dim}}
  \vspace{.2\baselineskip}
  }
  \\
  \hline %  ROW 3
\parbox{0.6cm}{ 23 } &
\parbox{0.6cm}{ 0.8} &
\parbox{10.0cm}{
  \vspace{.2\baselineskip}
  \textincon{ \small{C:7 F:maj$|$F:min$|$C:maj...C:maj$|$G:7$|$ C:maj}}
  \vspace{.1\baselineskip}
  }
  \\
  \hline %  ROW 4
\parbox{0.6cm}{ 23 } &
\parbox{0.6cm}{ 1.2} &
\parbox{10.0cm}{
  \vspace{.2\baselineskip}
  \textincon{\small{C:7/5 C:7 $|$ F:maj6 F\#:dim $|$ C:6(9) $|$ C:6(9) $|$ C:6(9) C:6(9) C:6(9) C:maj $|$ E:7(b9) $|$ A:min(6,9) A\#:min(6,9) $|$ A\#:min(6,9) $|$... D:min $|$ G:9 C:maj $|$ ...  G:7 \texttt{\char`_END\char`_} \texttt{\char`_START\char`_} C:maj }}
  \vspace{.2\baselineskip}
  }
  \\
\hline % LAST LINE
%\end{tabular}

%\end{center}
%\end{table}
\multicolumn{3}{c}{(a)} \\
% ---------------------%

%\begin{table}[h]
%\begin{center}
%\begin{tabular}{|c|c|c|}
\hline %  ROW 0
\parbox{0.6cm}{ 1 } &
\parbox{0.6cm}{ 0.5} &
\parbox{10.0cm}{
  \vspace{.2\baselineskip}
  \textincon{\small{C:maj $|$ G:7 $|$ ... $|$  G:7 $|$ C:maj6}}
  \vspace{.2\baselineskip}
  }
  \\
  \hline % ROW 2
\parbox{0.6cm}{ 1 } &
\parbox{0.6cm}{ 1.2} &
\parbox{10.0cm}{
  \vspace{.2\baselineskip}
  \textincon{\small{...C:maj \texttt{\char`_END\char`_} ... \texttt{\char`_START\char`_} \texttt{\char`_START\char`_} C\#:maj A\#:min A:sus4/5 C:maj/3 $|$ F:min7 A:min7 D:min7 D:min7... \texttt{\char`_START\char`_} }}
  \vspace{.2\baselineskip}
  }
  \\
  \hline %  ROW 3
\parbox{0.6cm}{ 8 } &
\parbox{0.6cm}{ 0.5} &
\parbox{10.0cm}{
  \vspace{.2\baselineskip}
  \textincon{ \small{C:maj A:min $|$ D:min7 G:7(b9)  $|$ C:maj $|$ A:min7  $|$ D:9 $|$ D:9 $|$ D:7 $|$ D:min7 $|$ G:7 $|$ C:maj $|$ C:7  $|$ F:maj  $|$  F:min  $|$ C:maj }}
  \vspace{.2\baselineskip}
  }
  \\
  \hline %  ROW 4
\parbox{0.6cm}{ 8 } &
\parbox{0.6cm}{ 1.2} &
\parbox{10.0cm}{
  \vspace{.2\baselineskip}
  \textincon{\small{C:Maj $|$ G:min7 $|$ F:maj $|$ D:min7 D:min7 D:min7/4 G:sus4(b7) $|$ G:min9 G:min9 G:min9 F\#:(1,3,b5,b7,9,13) $|$ C:6(9) G:sus4(b7,9) ... ... C:min \texttt{\char`_END\char`_} \texttt{\char`_START\char`_} C:maj }}
  \vspace{.2\baselineskip}
  }
  \\
\hline % LAST LINE
\end{tabular}

(b)
\caption{Chord progressions generated by char-RNN (a) and word-RNN (b). Bar symbols ($|$) are inserted for readability and repeated chords in each bar are omitted. }
\label{table:chord_results}	
\end{center}
\end{table}

% sufficient number of iterations => How many?
First, both char-RNN and word-RNN showed well-structured results. They learned the local structures of chords and bars after sufficient number of iterations. In the result, the majority of chords continued for multiples of four, implying a single chord for within a bar. They also learned the local relationships between flags and chord. After one iteration, the flags are not placed properly as in the table \ref{table:chord_results} (a), where \texttt{\char`_END\char`_} is not followed by \texttt{\char`_START\char`_} but repeats itself. As training continues, the flags start to appear in a sequence of \textincon{\texttt{\char`_END\char`_} \texttt{\char`_START\char`_} C:maj} as in the training texts. The last chords of the score, i.e., the chord before \texttt{\char`_END\char`_} are not always same as the first chord (C), which is also natural as they vary in the training file.  

Second, after sufficient training, both results showed chord progressions that lie in Jazz grammar. Examples are II-V-I progressions (D:min7- G:9 - C:maj), passing chords (A:dim - Ab:dim - G:min7), modal interchange chords (C:min6, Db:maj ) and substitutions (B:7 as a tritone subdominant of F:7) in char-RNN; modal interchanges (G:min7), circle of fifths (Eb:sus - Gb:maj6 - B:maj7), and descending bass (C:maj6,9 - B:dim - A:min7 - Ab:7) in word-RNN. The authors noticed a subtle difference between the results from the two approaches. The results from word-RNN are more conventional progressions than those of char-RNN. However, it cannot be the fundamental difference of the two approaches. Instead, it may be caused by the difference of effective lengths between char- and word-RNNs layers - they have the same length of state sequences, but it results in a longer chord sequence in the word-RNN as mentioned in Section \ref{sec:proposed}. In other words, the short memory of char-RNN may result in predictions that seem to be less constrained and stereotyped.

%------------------------------------------------------%
% \vspace{-0.5 cm}
% \vskip -0.5cm
%%%%%%%%%%%%%%%%%%%%%%%%%%%%%%%%
% CASE 2. DRUM TRACKS
%%%%%%%%%%%%%%%%%%%%%%%%%%%%%%%%
\section{Case 2. Drum Tracks}\label{sec:rhythm}
\subsection{Representation}
There are issues when applying LSTM networks to drum tracks including finding a way to create and effective text representation. Both chord progressions and drum tracks are sequences of simultaneous events (pitches and drum components). However, drum tracks do not have a meaningful and compressive representation such as chord and it necessitate an encoding strategy of the track into text. We also need a finer time resolution as generally there are more than four events in a bar.

To encode simultaneous events in a track into texts, we used a binary representation of \textit{pitches}, i.e., components of drums - kick, snare, hi-hats, cymbals, and tom-toms. For example, \texttt{100000000} and \texttt{010000000} represent kick and snare, respectively, and a simultaneous playing of kick and snare can be represented by \texttt{110000000}.%\footnote{Each binary code can be understood as a \textit{chord} of drum tracks. }

For efficient representation and learning, only nine components were allowed; kick, snare, open hi-hats, closed hi-hats, three tom-toms, crash cymbal, and ride cymbal.\footnote{Some of the components in the texts also represent other similar components, e.g. a closed hi-hats in the texts can mean either closed hi-hats or pedalled hi-hats in the original midi file.} We limited the number of events in a bar to $16$ by quantising the drum track by $16$th-note.

In the experiment, we first loaded $60$ midi files of drum tracks of \textit{Metallica} and quantised them. Then they were encoded into the above described binary representation. We also added a flag \texttt{\char`_BAR\char`_} as an annotation of the bar segments in order to check if the networks learns the local structure. 

There can be theoretically $2^9=512$ words, but there are supposedly much fewer words because the combinations of drum components that are played simultaneously are limited. The size of the word vocabulary in the training file is $119$ and the file consists of $2$,$141$,$692$ words in total.

\subsection{Results}

\begin{center}
\begin{figure*}[t]
\centering
  \includegraphics[width=0.9\textwidth]{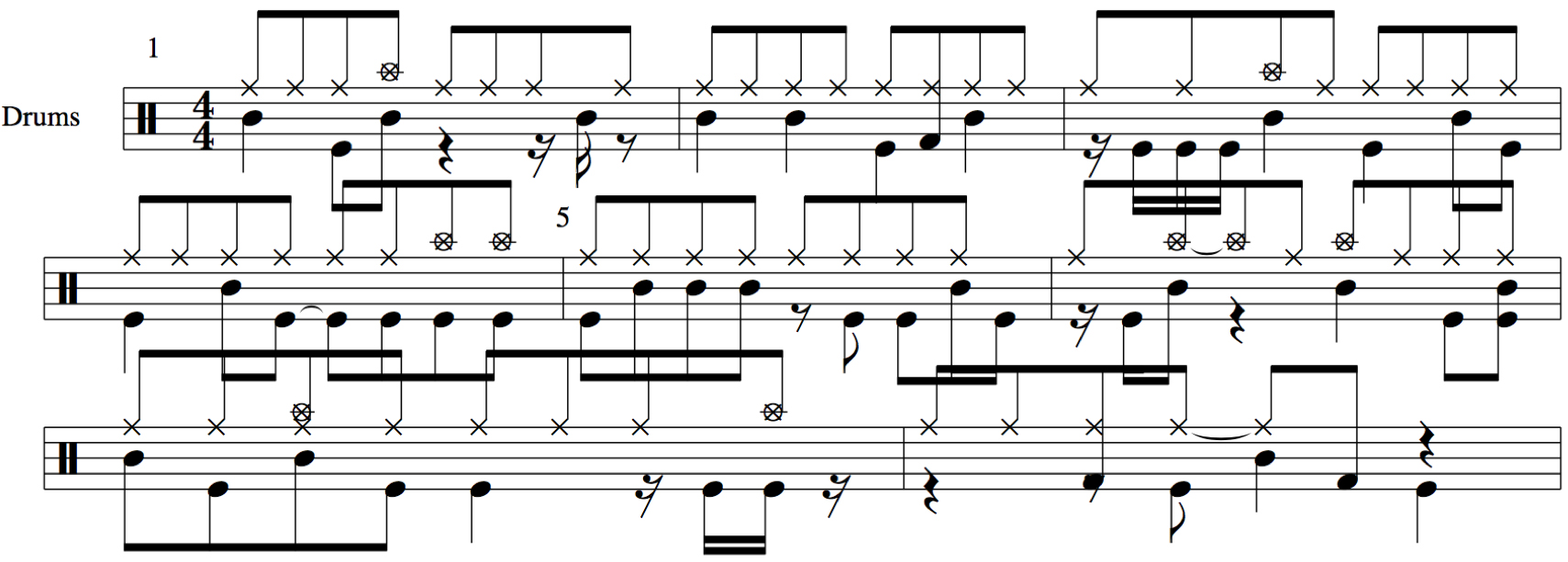}
  \caption{A score of a generated drum track. }\label{fig:drum_result}
\end{figure*}
\end{center}
\vspace{-0.7cm}
Char-RNNs turned out to fail to learn the drum tracks and output arbitrary $0$'s and $1$'s without any structures (the results have no spaces or \texttt{\char`_BAR\char`_} flags). The length of network may be too short to learn the long-term relationship between characters. In char-RNNs, representing a single bar requires $16$ events$\times$$10$ characters$=$$160$ time steps. Encoding music sequences with only two characters - $0$ and $1$ (+space to for segmentation) - is an extreme approach for char-RNNs. In this paper, we therefore only report the result of word-RNNs.

Figure \ref{fig:drum_result} shows one example of our results - a part of the generated track with $\alpha=1.0$ after $25$ iterations.\footnote{The score uses the percussion clef where $\times$ refers to hi-hats, notes on middle and bottom lines refers to snare and kick, respectively.} It consists of reasonable rock drum patterns - $8$-beat hi-hats, combinations of kick and snare, and occasional crash cymbals and tom-toms. Although there are occasional kick/snare/tom-toms notes on back beats (of sixteen notes), hi-hats remain consistent, playing on $4$-beat and $8$-beat pattern, which is very common for instance in drum tracks of Metallica.\footnote{\url{https://soundcloud.com/kchoi-research/00-24-100-bonus-for-score},\\The score in the figure starts from 34-second.}

Controlling $\alpha$ provides a way to tune the technical virtuosity of the track. Since large $\alpha$ increases the probabilities of occasional events, large $\alpha$ (=$1.5$) results in tracks with many fill-ins with tom-toms and a crash cymbal. On the other hands, when $\alpha < 1$, the track almost never contains anything but kick, snare, and hi-hats. As a result, it is possible to use a combination of small and large $\alpha$ in a drum track generator that is guided by user, who specifies where to add fill-ins.

%------------------------------------------------------%
% \section{Discussion}\label{sec:dis}

% \pagebreak

%------------------------------------------------------%
\section{Conclusion}\label{sec:con}
We introduced an algorithm of text-based LSTM networks for automatic composition and reported results for generating chord progressions and rock drum tracks. Word-RNNs showed good results in both cases while char-RNNs only successfully learned chord progressions. The experiments show LSTM provides a way to learn the sequence of musical events even when the data is given as text. With the diversity parameter, the proposed algorithm can be used as a tool that helps human composers. In the future, a more complex network with the capability of learning interactions within music (instruments, melody/lyrics) will be examined for a more complete automatic composition algorithm.

\bibliographystyle{splncs03}
%------------------------------------------------------%
\bibliography{csmc_bib}

\end{document}